\documentclass[letterpaper]{article} % DO NOT CHANGE THIS
\usepackage{aaai2026}  % DO NOT CHANGE THIS
\usepackage{times}  % DO NOT CHANGE THIS
\usepackage{helvet}  % DO NOT CHANGE THIS
\usepackage{courier}  % DO NOT CHANGE THIS
\usepackage[hyphens]{url}  % DO NOT CHANGE THIS
\usepackage{graphicx} % DO NOT CHANGE THIS
\usepackage{natbib}  % DO NOT CHANGE THIS AND DO NOT ADD ANY OPTIONS TO IT
\usepackage{caption} % DO NOT CHANGE THIS AND DO NOT ADD ANY OPTIONS TO IT
\usepackage{algorithm}
\usepackage{algorithmic}

\usepackage{newfloat}
\usepackage{listings}
\DeclareCaptionStyle{ruled}{labelfont=normalfont,labelsep=colon,strut=off} % DO NOT CHANGE THIS
\floatstyle{ruled}
\newfloat{listing}{tb}{lst}{}
\floatname{listing}{Listing}
\usepackage{pgfplots}
\pgfplotsset{compat=1.17}
\usepackage{rotating}
\usepackage{amssymb}
\usepackage{amsmath}
\usepackage{xcolor}
\usepackage{amsthm}
\usepackage{xcolor}
\usepackage{multirow}
\usepackage{makecell}
\newcounter{checksubsection}
\newcounter{checkitem}[checksubsection]

\usepackage{pgfplots}
\usepgfplotslibrary{groupplots}

\title{SVD-NO: Learning PDE Solution Operators with SVD Integral Kernels}

\author{
  Noam Koren\textsuperscript{\rm 1},
  Ralf J.\,J.\,Mackenbach\textsuperscript{\rm 2},
  Ruud J.\,G.\,van~Sloun\textsuperscript{\rm 3},
  Kira Radinsky\textsuperscript{\rm 1},
  Daniel Freedman\textsuperscript{\rm 4}
}
\affiliations{
  \textsuperscript{\rm 1}Department of Computer Science, Technion -- Israel Institute of Technology, Haifa, Israel\\
  \textsuperscript{\rm 2}Swiss Plasma Center, EPFL, Switzerland\\
  \textsuperscript{\rm 3}Department of Electrical Engineering, Eindhoven University of Technology, Eindhoven, The Netherlands\\
  \textsuperscript{\rm 4}Department of Applied Mathematics, Tel Aviv University, Tel Aviv, Israel\\
  noam.koren@campus.technion.ac.il,\ ralf.mackenbach@epfl.ch,\ r.j.g.v.sloun@tue.nl,\\
  kira@technion.ac.il,\ dfreedman@tauex.tau.ac.il
}

\begin{document}

\maketitle
\begin{abstract}
Neural operators have emerged as a promising paradigm for learning solution operators of partial differential equations (PDEs) directly from data.  Existing methods, such as those based on Fourier or graph techniques, make strong assumptions about the structure of the kernel integral operator, assumptions which may limit expressivity.  We present SVD-NO, a neural operator that explicitly parameterizes the kernel by its singular-value decomposition (SVD) and then carries out the integral directly in the low-rank basis. Two lightweight networks learn the left and right singular functions, a diagonal parameter matrix learns the singular values, and a Gram-matrix regularizer enforces orthonormality.  As SVD-NO approximates the full kernel, it obtains a high degree of expressivity.  Furthermore, due to its low-rank structure the computational complexity of applying the operator remains reasonable, leading to a practical system.  In extensive evaluations on five diverse benchmark equations, SVD-NO achieves a new state of the art.  In particular, SVD-NO provides greater performance gains on PDEs whose solutions are highly spatially variable.\\
The code of this work is publicly available at \footnote{https://github.com/2noamk/SVDNO.git}.
\end{abstract}

\section{Introduction}
Many problems in physics and engineering can be modeled by Partial Differential Equations (PDEs), which describe the evolution of complex systems across space and time. While classical numerical solvers, such as the finite element method \cite{fem}, finite difference method \cite{fdm}, and finite volume method \cite{fvm}, are widely used, they often require fine spatial discretization and massive computational resources. For example, simulating turbulent plasma dynamics can demand millions of CPU hours \cite{tang2017global}. 
These constraints have encouraged the rise of various techniques based on deep learning, from data-driven surrogates such as U-Net models \cite{gupta2022towards, unet}, to neural operators that learn the entire solution operator from data and now represent the state of the art.

Neural operators approximate functionals between infinite-dimensional function spaces.  The functionals map from some specification of the PDE, such as initial conditions, boundary conditions, or parameter fields, to the solution of the PDE.  Neural operators can be broadly divided into four established families.
\textbf{DeepONet models} \cite{deeponet} break down the operator learning problem into one of learning two subnetworks, the branch which processes the input function and the trunk which processes the query point.  Overall, DeepONets tend to be less effective than methods which parameterize the integral operator directly; the latter are represented by the following three families.
\textbf{Fourier-based operators} \cite{fno} recast the kernel integral as a convolution evaluated in the frequency domain.  This implies both stationarity of the operator as well as a lack of dependence of the operator on the input function; both of these implications limit the expressivity of the operator.
\textbf{Graph-based operators} \cite{gkn} employ message passing on both regular grids as well as irregular meshes, which enables them to accommodate complex geometries.  As the message passing is local in nature, the underlying kernel integral operator must also be local.  In practice, this means that such methods often oversmooth and struggle to capture long-range interactions without computational overhead.
\textbf{Physics-informed operators} \cite{li2024physics} take inspiration from Physics-Informed Neural Networks \cite{pinn}, by including a loss based on the residual of the PDE.  Such a loss can often be useful, but does not directly address the expressivity of the underlying kernel integral operator.  (For example, in \cite{li2024physics} the method used was a Fourier Neural Operator.)

In this paper, we introduce the \textbf{Singular Neural Operator (SVD-NO)}, a method that is derived from the underlying functional analysis of integral operators. SVD-NO begins with a full kernel integral operator, and then expresses it in a useful form via its singular value decomposition.  By truncating this decomposition, SVD-NO is able to \textit{directly} learn the operator.  Specifically, it learns both the kernel’s left and right singular functions along with their singular values.  The only approximation is the truncation, which imposes a low-rank structure.  As a result, the operator retains a high degree of expressivity, as it may depend in complex ways on both the coordinates as well as the input function.  Unlike Fourier-based methods, there are no stationarity assumptions; and unlike graph-based based methods, the kernel is not assumed to be local.  Finally, due to the structure of the SVD decomposition, the application of the integral operator to a given function is rendered efficient, allowing the method to be practical.

In summary, the contributions of this work are threefold:  \\
1. \textbf{Utilization of operator theory to motivate a new architecture for solving PDEs.} We embed a learnable SVD decomposition of the Hilbert–Schmidt kernel within a neural operator. Although the mathematics of the singular value decomposition of integral kernels has classical roots, this is the first time it has been realized as an end-to-end trainable layer, yielding a low-rank parameterization grounded in functional analysis.\\
2. \textbf{An expressive yet practical neural operator based on SVD.} Other than assuming a low-rank structure, our method makes no other approximations or assumptions on the form of the integral kernel.  The result is a high degree of expressivity: our architecture allows for a highly complex dependence on the input function, by modelling an approximation to the full kernel operator.  This expressivity is considerably greater than competing methods such as Fourier- or graph-based techniques, which make strong assumptions on the form of the operator.  Furthermore, due to the low-rank structure, the computational complexity of applying the operator remains reasonable, leading to a practical system. \\
3. \textbf{State-of-the-art performance on diverse PDEs.} The increased expressivity of SVD-NO leads to strong empirical results.  We evaluate the proposed operator on five benchmark equations and show significant improvements over six leading neural operators.

\section{Related Work}

\subsection{Survey of Neural Operators}

\label{related_work}
Neural operators learn mappings between infinite-dimensional function spaces, from some specification of the PDE - such as initial conditions, boundary conditions, or parameter fields - to the solution of the PDE.  Thus, they learn the \emph{solution operator} of PDEs. Broadly, they can be grouped into four major categories, each characterized by how they approximate the underlying \emph{kernel integral operator}: DeepONet-based, Fourier-based, Graph-based, and Physics-Informed.

\paragraph{DeepONet-Based Operators.}
DeepONet~\cite{deeponet} was among the first neural operator. It learns operator mappings directly from data without discretizing the kernel. 
It introduced a branch–trunk architecture: the branch network embeds sampled values of the input field, the trunk encodes the query point, and their inner product yields the solution.
Follow-ups, e.g., F-DeepONet~\cite{fdeeponet}, PI-DeepONet~\cite{olpinn}, and GraphDeepONet~\cite{graphdeeponet}, add Fourier features, physics-informed losses, or graph message passing, respectively. 
Despite these advances, DeepONet often struggles with high-dimensional inputs due to its fully connected structure.

\paragraph{Fourier-Based Neural Operators.}
The Fourier Neural Operator (FNO)~\cite{fno} treats the kernel as a convolution and evaluates it in the frequency domain via the Fourier transform.  This assumes that the solution operator is stationary.  Several variants have been proposed:
F-FNO~\cite{ffno} applies low-rank factorizations to the Fourier weight matrices to reduce computational complexity.
WNO~\cite{waveletno} replaces Fourier transforms with wavelets for better spatial localization.
SFNO~\cite{sfno} and Geo-FNO~\cite{geofno} adapt FNO to spherical and irregular domains, respectively.
U-NO~\cite{uno} integrates FNO into a U-Net encoder-decoder structure for multiscale modeling.

\paragraph{Graph-Based Neural Operators.} 
For PDEs on unstructured domains, Graph Neural Networks (GNNs)~\cite{gnn} are a natural representation. The Graph Neural Operator (GNO)~\cite{gkn} models functions over nodes and approximates the kernel by message passing on graph edges. While effective on irregular geometries, these methods often struggle capturing long-range dependencies and can suffer from oversmoothing and high computational cost. 
Several extensions were researched; LGN~\cite{lgn} simulates 3D lattice structures. GINO~\cite{gino} bridges GNO and FNO by projecting irregular grids to regular ones for FFT-based modeling.
The Message-Passing Neural PDE Solver (MPNN)~\cite{mpnn} approximates the operator by passing messages along graph edges. 

\paragraph{Physics-Informed Neural Operators.} PINO methods incorporate the structure of the PDE directly into the loss function, enabling data-efficient learning guided by physical laws. For example, \cite{pino} adds physics-informed loss terms to guide the FNO training with PDE residuals.  This approach is somewhat orthogonal, in that this loss may be used for any given neural operator architecture.

\subsection{Relationship with SVD-NO}

All approaches approximate the kernel operator indirectly, either by learning it functionally via neural networks (e.g., DeepONet, GNN-based operators, PINNs) or by performing spectral analysis using Fourier transforms under stationarity assumptions, with Fourier-based neural operators (e.g. FNO, U-NO) and physics-informed variants (e.g. PINO) currently leading in accuracy.  
Addressing this gap, SVD-NO explicitly learns the kernel and applies the operator by integrating the learned kernel against the input field, aligned with classical operator theory.

We benchmark SVD–NO against the six established models from the PDENNEval benchmark \cite{pdenneval}, a comprehensive evaluation suite for neural-network PDE solvers: DeepONet, FNO, U-NO, MPNN, PINO, and U-Net. Together, these models span the four operator-learning families discussed above, while U-Net serves as an ML baseline.

% \subsection{Singular Value Decomposition (SVD)}
% Singular Value Decomposition (SVD) for functions generalizes the classical matrix SVD \cite{svd} to function spaces. Given a collection of functions or an operator defined on \(L^2(\Omega)\), the SVD expresses it as a sum of orthonormal basis functions weighted by singular values:
% \[
% f(x, y) \approx \sum_{\ell=1}^{L} \sigma_\ell\, \phi_\ell(x)\, \psi_\ell(y),
% \]
% where \(\{\phi_\ell\}\) and \(\{\psi_\ell\}\) are orthonormal function sets in \(L^2(\Omega')\) and \(L^2(\Omega)\), respectively, and \(\{\sigma_\ell\}\) are non-negative singular values ordered in decreasing magnitude. This decomposition enables efficient approximation, compression, and analysis of functional data and serves as the foundation for low-rank representations. In our work, we leverage this framework to construct the SVD neural operator by directly learning the singular components of the solution kernel.

\section{Neural Operator Background}
\label{sec:neural_ops}

\paragraph{Operator Learning.}
Let $D \subset \mathbb{R}^{d_x}$ be a bounded domain and consider two Hilbert spaces of functions $\mathcal A:=C\!\bigl(D;\,\mathbb{R}^{d_a}\bigr)$ and
$\mathcal U:=C\!\bigl(D;\,\mathbb{R}^{d_u}\bigr)$.
%equipped with norms $\|\cdot\|_{\mathcal{A}}$ and $\|\cdot\|_{\mathcal U}$, respectively.
An operator
\[
\mathcal G:\mathcal A\longrightarrow\mathcal U
\]
maps an \emph{input field} $a\in\mathcal A$ (e.g.\ coefficients, source
terms, initial or boundary data) to a corresponding \emph{output field}
$u=\mathcal G(a)\in\mathcal U$ (e.g.\ the PDE solution).
Learning~$\mathcal G$ from a finite set of
samples $\{(a_j,u_j)\}_{j=1}^{N}$ with $u_j=\mathcal G(a_j)$, where $N$ is the number of samples,
is the essence of \emph{operator learning}
\citep{Neuralop} and forms the foundation of neural
operators.

\vspace{0.5em}
\paragraph{Neural Operator Architecture.}
A neural operator parameterizes an \emph{approximate} operator
\(
\mathcal G_\theta:\mathcal A\to\mathcal U
\)
through the composition
\begin{equation}
\begin{aligned}
\mathcal G_\theta \;=\;
Q\circ
\bigl[\gamma\,(W^{T-1}+\mathcal{K}_\theta^{T-1})\bigr]\circ
\cdots 
\\
\circ
\bigl[\gamma\,(W^{0}+\mathcal{K}_\theta^{0})\bigr]\circ
P
\label{eq:no_arch}
\end{aligned}
\end{equation}
where
\begin{itemize}
\item
$P:\mathbb R^{d_a}\!\to\!\mathbb R^{d_v}$ and
$Q:\mathbb R^{d_v}\!\to\!\mathbb R^{d_u}$ are \emph{local} fully
connected networks that \emph{lift} the input
$a(x)$ to a high-dimensional latent field
$v^{0}(x)=P\!\bigl(a(x)\bigr)$; and \emph{project} the final latent
representation back to the target space $u(x) = Q(v^T(x))$, where $v^T$ is the last hidden layer.
\item
Each hidden layer applies a pointwise linear map
$W^t:\mathbb R^{d_v}\!\to\!\mathbb R^{d_v}$ in parallel with a
\emph{non-local integral operator}
\(
\mathcal{K}_\theta^{t}(a):\mathcal U\!\to\!\mathcal U,
\)
followed by a non-linearity~$\gamma$:

\begin{equation}
\begin{aligned}
v^{t+1}(x) &=
  \gamma\!\Bigl(
    W^{t}v^{t}(x)
    + [\mathcal{K}_{\theta}^{t}(a)\,v^{t}](x)
  \Bigr)
\end{aligned}
\label{eq:no_layer}
\end{equation}
\item
The integral operator $\mathcal{K}_\theta^{t}(a)$ is defined by a \emph{kernel}, $\kappa_\theta^{t}$:
\begin{equation}
\begin{aligned}
\bigl[\mathcal{K}_\theta^{t}(a)\,v\bigr](x)
=
\int_{D}
\kappa_\theta^{t}\!\bigl(x, a(x),x', a(x')\bigr)\;v(x')\,\mathrm dx'
\label{eq:no_kernel}
\end{aligned}
\end{equation}
\end{itemize}
\vspace{0.5em}
\paragraph{Training.}
Given training pairs $\{(a_j,u_j)\}_{j=1}^{N}$ we learn~$\theta$ by
minimising an empirical risk,
\(
\min_{\theta}\frac1N\sum_{j=1}^{N}
\mathcal L\!\bigl(\mathcal G_{\theta}(a_j),u_j\bigr),
\)
where $\mathcal L$ is typically an $L_{2}$ or relative-error loss.

% \begin{equation}
% \begin{aligned}
% \mathcal{L}_2 \;=\;
%     \frac{1}{N}\,
%     \sum_{i=1}^{N}
%     \frac{\;\lVert u_i - \hat{u}_i \rVert_{2}\;}
%          {\lVert u_i \rVert_{2}},\quad u_i,\hat{u}_i\in\mathbb{R}^{n},
%     \label{eq:l2}
% \end{aligned}
% \end{equation}

\begin{figure*}[t]
\centering
\includegraphics[width=.65\textwidth]{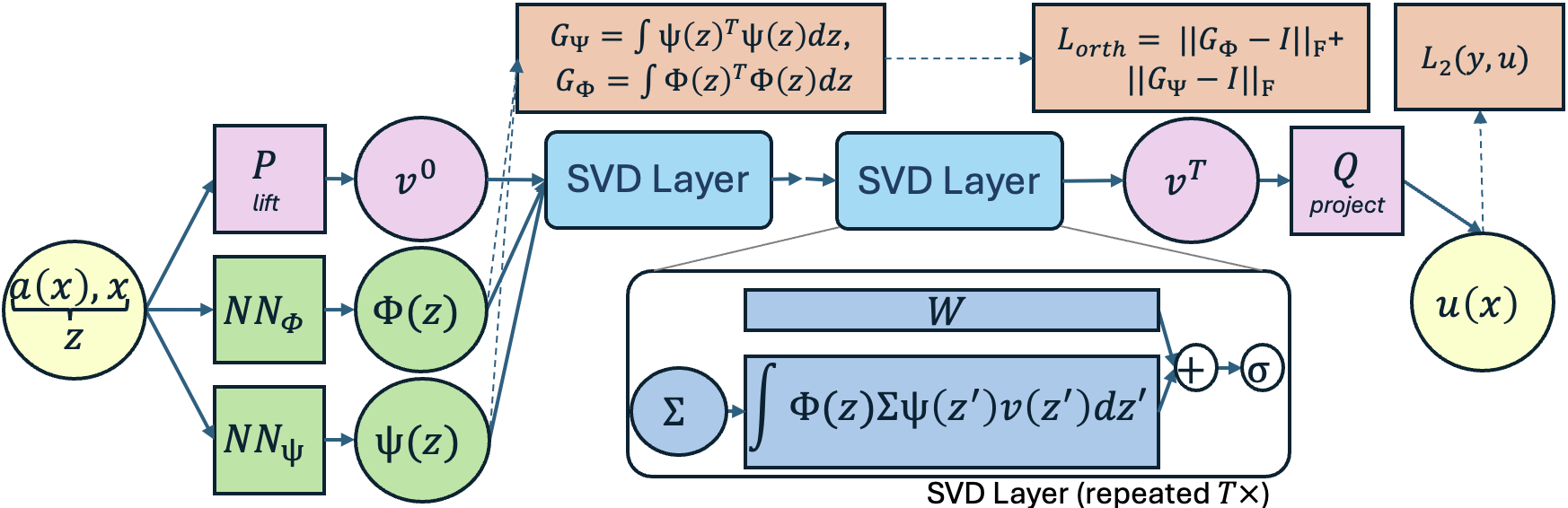}
\caption{\textbf{SVD-NO.}
The coefficient–coordinate tuple \(z=(a(x),x)\) is fed through two singular function nets, \(\mathrm{NN}_{\!\Phi}\) and
\(\mathrm{NN}_{\!\Psi}\), producing global basis functions \(\Phi(z)\) and
\(\Psi(z)\). An encoder \(P\) maps \(z\) to the initial latent state, 
\(v^{0}\), which is then processed by \(T\)
\emph{SVD blocks} to reach \(v^{T}\).
Each block applies a kernel
\(K(v)=\Psi(z)\,\Sigma\!\int\!\Phi(z')\,v(z')\,dz'\)
(with trainable diagonal \(\Sigma\)), adds a point-wise map \(Wv\),
followed by activation~\(\sigma\).
A decoder \(Q\) maps the final latent state, \(v^T\), to the predicted solution \(u\).
Dashed arrows indicate loss paths: Gram matrices
\(G_{\!\Phi},G_{\!\Psi}\) drive the orthogonality penalty
\(L_{\text{orth}}\), with \(u\) supervising \(L_{2}(y,u)\).}
\label{fig:svd_no_figure}
\end{figure*}

\section{The Singular Neural Operator (SVD-NO)}
\label{sec:SVD-NO}

\subsection{SVD-NO Theory and Architecture}

\paragraph{Integral Layer Reformulation.}
Let $D\subset\mathbb{R}$ be a bounded spatial domain and let $a\colon D\to\mathbb{R}^{d_a}$ denote an input field, recall the generic integral layer
\vspace{-0.3em}
\[
\bigl[\mathcal K_\theta(a)\,v\bigr](x)
\;=\;
\int_{D}
\kappa_\theta\!\bigl(x, a(x), x', a(x')\bigr)\,
v(y)\,\mathrm dy
\quad (\text{Eq.\,\ref{eq:no_kernel}}),
\]
Henceforward we will suppress $\theta$. It is convenient to regard the pair $(x,a(x))$ as a single augmented coordinate
\begin{equation}
    z \;=\; (x,a(x)) \in \mathcal{Z} \;=\; D \times A,
\end{equation}
where $A \subset \mathbb{R}^{d_a}$ is the range of the input function $a(\cdot)$. Using this notation the kernel becomes a mapping
$\kappa \colon \mathcal{Z}\times\mathcal{Z}\to\mathbb{R}^{d \times d}$, and Eq.~\eqref{eq:no_kernel} reads
\begin{equation}
    (\mathcal{K}v^{t})(z)
    \;=\;
    \int_{\mathcal{Z}} \kappa(z,z')\,v^{t}(z')\,\mathrm{d}\mu(z'),
    \qquad z\in\mathcal{Z},
    \label{eq:operator_augmented}
\end{equation}
where $\mu$ is the push‑forward of the Lebesgue measure on~$D$ under $x\mapsto (x,a(x))$.

SVD-NO retains this structure but \emph{constrains} the kernel $\kappa_\theta$ through an explicit low-rank singular-value decomposition (SVD). Figure \ref{fig:svd_no_figure} provides an overview.

\paragraph{An SVD Parameterization: Scalar Case.}
\label{sec:svd_approach}

Suppose that each of the $v^t$ functions are scalar-valued (i.e.~choose $d=1$), so that $\kappa$ is also scalar-valued.  If we choose the function $\kappa$ so that it is square integrable, then the integral operator  
\[
(\mathcal Kv)(z)
\]
is Hilbert–Schmidt and therefore compact.  Consequently there exist orthonormal
systems $\{\phi_\ell\}_{\ell \in \mathbb{N}}$ and $\{\psi_\ell\}_{\ell \in \mathbb{N}}$ in $L_\mu^2(\mathcal Z)$ together with singular values
$\{\sigma_\ell\}_{\ell \in \mathbb{N}}\!\in\!\ell^{2}$ such that
\begin{equation}
\label{eq:svd}
\kappa(z,z') \;=\;
\sum_{\ell=1}^{\infty} \sigma_\ell\,\phi_\ell(z)\,\psi_\ell(z'),
\qquad\forall\,z,z'\in\mathcal Z,
\end{equation}
with absolute convergence \cite{arcozzi2025functional, rudin1991functional}.  A proof of this statement is given in the Appendix.  In practice we retain only the leading
$L$ terms of~\eqref{eq:svd}, yielding the rank-$L$ approximation
$\kappa_L(z,z')=\sum_{\ell=1}^{L}\sigma_\ell\phi_\ell(z)\psi_\ell(z')$.

\paragraph{Update Equation.}
Inserting \eqref{eq:svd} into the evolution step of a Neural Operator gives
\begin{equation}
\begin{aligned}
(\mathcal Kv)(z)\;=\;\!\int_{\mathcal Z}\sum_{\ell=1}^{L}\sigma_\ell\phi_\ell(z)\psi_\ell(z')\,v(z')\,\mathrm d\mu(z')
\end{aligned}
\end{equation}
which simplifies to
\begin{equation}
\label{eq:update_continuous}
v^{t+1}(z) \;=\;
\sum_{\ell=1}^{L} \sigma_\ell\,\phi_\ell(z)\!
\int_{\mathcal Z} \psi_\ell(z')\,v^{t}(z')\,\mathrm d\mu(z').
\end{equation}
The ability to factor out the $z$ part of the expression from the integral will have important implications for the computational complexity, as we will see in Section \ref{sec:SVDNO_complexity}.

\paragraph{Extension to Vector-Valued Fields.}
We may improve the expressivity of the neural operator by allowing the $v^t$ functions to be vector-valued, i.e.~$v^t:\mathcal Z\!\to\!\mathbb R^{d}$.  There is a natural extension of the SVD decomposition for this scenario, which is as follows.  We promote the kernel to
$\kappa:\mathcal Z\times\mathcal Z\!\to\!\mathbb R^{d\times d}$ and write
\[
\kappa(z,z') \;=\; \Phi(z)\,\Sigma\,\Psi(z')^{\!\top},
\]
where
$\Phi(z),\Psi(z')\in\mathbb R^{d\times L}$ collect the $d$-component singular functions
$\{\phi_\ell(z)\}_{\ell=1}^{L}$ and $\{\psi_\ell(z')\}_{\ell=1}^{L}$, and
$\Sigma=\mathrm{diag}(\sigma_1,\dots,\sigma_L)$ is a diagonal matrix containing the singular values.  In this case, we have that
\[
(\mathcal Kv)(z)\;=\;\!\int_{\mathcal Z}\Phi(z)\,\Sigma\,\Psi(z')^{\!\top}\,v(z')\,\mathrm d(z')
\]
which simplifies to
\[
(\mathcal Kv)(z)\;=\;\!\Phi(z)\,\Sigma\,\int_{\mathcal Z}\Psi(z')^{\!\top}\,v(z')\,\mathrm d(z')
\]

This extension is quite natural.  Although we have not proven its convergence as $L \to \infty$, a precisely analogous result for matrix-valued operators has been proven in the case of positive definite operators, which is effectively a matrix version of Mercer's Theorem \cite{mercer_extension}.  (We also note a different setting for functional SVD has been treated in \cite{tan2024functional}.)
%SVD-NO drops the symmetry requirement and learns the low-rank factors \(\Phi(z),\Psi(z')\) and singular values \(\Sigma\).

\paragraph{Discrete Form.}
With quadrature nodes $\{z_j\}_{j=1}^{n}$ and weights $\{\Delta z_j\}$,
the integral operator is approximated by
\begin{equation}
\begin{aligned}
\label{eq:update_discrete}
v^{t+1}(z_i)
\;=\;
\Phi(z_i)\,\Sigma\,
\sum_{j=1}^{n}
\Psi(z_j)^{\!\top} v^{t}(z_j)\,
\Delta z
\end{aligned}
\end{equation}

\paragraph{Direct Factor Parameterization.}
Rather than first estimating $\kappa$ and then factorizing, we \emph{parameterize} $\kappa$ \emph{directly} by its SVD factors. Specifically, we introduce two singular function nets
\[
\Phi(\cdot;\theta_\phi),\;
\Psi(\cdot;\theta_\psi):\mathcal Z \;\longrightarrow\; \mathbb R^{d\times L},
\]
each realized by a lightweight neural network;
% (see Section~\ref{backbone} for details),
and a diagonal matrix of singular values
\[
\Sigma=\mathrm{diag}(\sigma_1,\dots,\sigma_L)\in\mathbb R^{L\times L},
\]
whose entries $\{\sigma_\ell\}$ are learnable parameters.  
The complete trainable set is therefore
\(
\theta=(\theta_\phi,\theta_\psi,\Sigma),
\)
optimised jointly via back-propagation.

\paragraph{Singular Function Nets (\(\Phi,\Psi\)).}
\label{backbone}
Each singular function net is implemented with one of the following lightweight neural networks:
\begin{itemize}
    \item \textbf{MLP}: This is effectively a neural field \cite{xie2022neural}.  It consists of linear layers with sine activations.
    \item \textbf{LSTM (1D only)}: LSTM layers (sigmoid gates, tanh outputs) with a linear projection. Effective at capturing long-range dependencies on 1D spatial grids, but not extensible to \(d_x>1\) as there is no natural ordering of points.
    % stacked LSTM layers (sigmoid gates, tanh output) followed by a linear projection.  
    % LSTMs are effective at capturing long-range dependencies in 1D sequential data, which is analogous to spatial grids in one dimension. 
    % LSTMs are limited to 1-D domains: when \(d>1\) there is no canonical ordering of spatial points, so sequence models become sensitive to arbitrary grid permutations and non-robust under re-meshing.
\end{itemize}

\paragraph{Orthogonality.}
\label{para:ortho}
The true singular functions form an orthonormal system,
\[
\langle\phi_\ell,\phi_k\rangle \;=\; \delta_{\ell k},
\qquad
\langle\psi_\ell,\psi_k\rangle \;=\; \delta_{\ell k},
\]
where $\delta_{\ell k}$ denotes the Kronecker delta.
To push the learned singular functions toward the same structure we introduce an
\emph{orthogonality penalty}.

The Gram matrices of $\Phi(z)$ and $\Psi(z)$ are:
\[
G_\Phi \;=\;
\int_{D}\!\Phi(z)^{\!\top}\Phi(z)\,dz,
\qquad
G_\Psi \;=\;
\int_{D}\!\Psi(z)^{\!\top}\Psi(z)\,dz,
\]
with integrals approximated via the trapezoidal rule.
Departures from perfect orthonormality are measured by the squared Frobenius distance to the identity:
\begin{equation}
\begin{aligned}
\mathcal{L}_{\text{ortho}}
\;=\;
\bigl\|G_\Phi - I_L\bigr\|_{F}^{2}
\;+\;
\bigl\|G_\Psi - I_L\bigr\|_{F}^{2},
    \label{eq:orth_loss}
\end{aligned}
\end{equation}
where $I_L$ is the $L\times L$ identity matrix.
Including $\mathcal{L}_{\text{ortho}}$ in the total loss softly regularizes
the columns of $\Phi$ and $\Psi$, nudging them toward an orthonormal basis.

\subsection{Expressivity and Computational Complexity}
\label{sec:SVDNO_complexity}

\paragraph{Improved Expressivity vs. Competing Methods.}  As we have seen, SVD-NO models the integral kernel $\kappa$ in such a way so as to allow full dependence on all arguments.  That is, we retain the property that $\kappa = \kappa(x, a(x), x', a(x'))$.  There are two important aspects to highlight here:
\begin{enumerate}
    \item \textbf{Input Function Dependence:} $\kappa$ depends not only on the coordinates $x$ and $x'$, but also on the input function through $a(x)$ and $a(x')$.
    \item \textbf{Long-Range Effects:} For a given $x$, $\kappa$ can take non-zero values, indeed even very large values, for $x'$ which is far away from $x$.
\end{enumerate}
Let us contrast this situation with competing methods.  In the case of Fourier-based methods, the kernel does not have input function dependence; the assumption is that it only depends on the coordinates, i.e.~$\kappa = \kappa(x, x')$.  Indeed the assumption is even stronger than that, in that it is only dependent on the \textit{difference} between coordinates, $\kappa = \kappa(x-x')$.  This limits the expressivity of the Fourier-based kernels.

In the case of graph-based methods, the underlying graph is taken to be \textit{local} in nature: a grid point only interacts with nearby points.  In other words, $\kappa = \kappa(x, a(x), x', a(x'))$ but is only non-zero for $x' \in \mathcal{N}(x)$, i.e.~in a small neighborhood around $x$.  Thus, in the case of graph-based methods the kernel cannot directly model long-range effects; instead, these must come from the repeated application of many layers, which can in practice lead to oversmoothing.  In Section \ref{sec:results}, we see that this greater expressivity improves performance.

\paragraph{Time Complexity.}
The complexity is governed by the discrete update rule in Eq.~\eqref{eq:update_discrete}
where $n$ is the number of evaluation points $z_i \in \mathcal{Z}$, $d$ is the dimensionality of the vector field $v^t(z) \in \mathbb{R}^d$, and $L$ is the number of retained SVD modes.

This update can be decomposed into two steps: \\
    (1) Compute rank-$L$ representation:
    \[
    q := \sum_{j=1}^{n} \Psi(z_j)^{\!\top} v^{t}(z_j)\, \Delta z \in \mathbb{R}^{L}.
    \]
    For each $z_j$, evaluating $\Psi(z_j)^{\!\top} v^{t}(z_j)$ requires $O(dL)$ operations. Summing over all $n$ points costs $O(n d L)$. \\
    (2) For each output point $z_i$:
    \[
    v^{t+1}(z_i) = \Phi(z_i)\, \Sigma\, q.
    \]
    Multiplying $\Sigma$ (diagonal matrix) with $q$ costs $O(L)$, and the matrix-vector product with $\Phi(z_i) \in \mathbb{R}^{d \times L}$ costs $O(d L)$. Repeating this for all $n$ points costs $O(n d L)$. \\
% \begin{itemize}
%     \item Compute rank-$L$ representation:
%     \[
%     q := \sum_{j=1}^{n} \Psi(z_j)^{\!\top} v^{t}(z_j)\, \Delta z \in \mathbb{R}^{L}.
%     \]
%     For each $z_j$, evaluating $\Psi(z_j)^{\!\top} v^{t}(z_j)$ requires $O(dL)$ operations. Summing over all $n$ points costs $O(n d L)$.
    
%     \item For each output point $z_i$:
%     \[
%     v^{t+1}(z_i) = \Phi(z_i)\, \Sigma\, q.
%     \]
%     Multiplying $\Sigma$ (diagonal matrix) with $q$ costs $O(L)$, and the matrix-vector product with $\Phi(z_i) \in \mathbb{R}^{d \times L}$ costs $O(d L)$. Repeating this for all $n$ points costs $O(n d L)$.
% \end{itemize}
Thus, the total complexity of one layer is $O(ndL)$.  This scales linearly with the number of points, output dimension, and SVD rank $L$.  This is a major improvement over the cost of generic kernel integral operators, which is easily seen to be $O(n^2 d^2)$, as $L \ll nd$.  (See the ablation in Section \ref{sec:ablation} for runtime comparisons.)

\paragraph{Memory Complexity.} Each SVD-NO layer must store:
\begin{itemize}
  \item The latent field \(v\in\mathbb{R}^{n\times d}\), costing \(O(nd)\) memory.
  \item The singular function net activations \(\Psi, \Phi\in\mathbb{R}^{n\times d\times L}\) costing \(O(ndL)\) memory; and $\Sigma$ costing $O(L)$.
\end{itemize}
Hence, the total peak memory footprint per layer is
\(O(ndL),
\)
i.e.\ linear in the number of points \(n\), feature dimension \(d\), and SVD rank \(L\). As in the case of time complexity, this improves substantially the \(O(n^2 d^2)\) memory cost of generic kernel integral operators.
(See Figure~\ref{fig:loss-and-mem} for memory usage.)

% \subsection{Benefits.}
% \label{sec:SVD-NO_benefits}

% The SVD-based parameterization of the kernel operator in SVD-NO offers two primary benefits, both in terms of expressiveness and efficiency:

% \begin{enumerate}
%     \item \textbf{Improved expressiveness over Fourier-based methods.} Unlike the Fourier-based methods, which implicitly assume stationarity, SVD-NO imposes no such structural constraint. Moreover, the learned modes $\phi_\ell(z)$ and $\psi_\ell(z)$ directly depend on the input field $a(\cdot)$, enabling a natural integration of input-dependent structures such as initial and boundary conditions.
%     (see Figure \ref{fig:spatial-stationarity} for SVD‐NO improvement across PDEs with varying stationarity).
   
%     \item \textbf{Reduced Complexity.} Compared to the general kernel integral operator in Eq.~\eqref{eq:no_kernel}, which requires $O(n^2 d^2)$ memory and time for $n$ spatial points and $d$ output dimensions, the SVD-based formulation reduces both to $O(n\,d\,L)$, where $L$ is the number of retained singular modes. Since $L \ll n$ in practice, this yields substantial speedups and lower memory while maintaining accuracy. 
% (see Ablation \textit{Direct MLP Kernel} for runtime comparisons, and Figure~\ref{fig:loss-and-mem} for memory usage).
% \end{enumerate}

\section{Empirical Evaluation}
\subsection{Experimental Methodology}
\label{sec:exp_method}
Our experiments are designed to quantify how well SVD-NO generalizes across five diverse benchmark equations: Shallow-Water, Allen-Cahn, Diffusion-Reaction, Diffusion-Sorption, and Darcy-Flow. We compare our results with SOTA baselines.  Performance is reported as the \emph{mean L\textsubscript{2} relative error}
multiplied by $100$:
\begin{equation}
\begin{aligned}
 L_2 = 100 \times \;
    \frac{1}{N}\,
    \sum_{i=1}^{N}
    \frac{\;\lVert u_i - \hat{u}_i \rVert_{2}\;}
         {\lVert u_i \rVert_{2}}
    \label{eq:mean_l2_rel_err}
\end{aligned}
\end{equation}
where $\lVert\cdot\rVert_{2}$ is the Euclidean norm.
This metric expresses the error as a percentage of the ground-truth signal
energy and is standard in operator-learning benchmarks
\cite{tripura2022wavelet}.
\subsection{Baseline Models}

We evaluated SVD-NO against the six representative baselines evaluated in the PDENNEval benchmark \cite{pdenneval}, a comprehensive evaluation suite for neural network PDE solvers: DeepONet, FNO, U-NO, MPNN, PINO, and U-Net. These models cover all four neural-operator families reviewed in Section~\ref{related_work} and include an ML surrogate (U-Net).

% \begin{table*}[ht!]
%     \scriptsize
%     \centering
%     \begin{tabular}{|c|c|c|c|c|c|}
%         \hline
%         Name & Equation & Variables & Input & Output & Sampling \\
%         \hline
%         2D Shallow Water
%         & \shortstack{
%             $\partial_t h + \partial_x(hu) + \partial_y(hv) = 0$ \\
%             $\partial_t (hu) + \partial_x \left( u^2 h + \frac{1}{2} g_r h^2 \right) + g_r h\, \partial_x b = 0$
%             \\ $\partial_t (hv) + \partial_y \left( v^2 h + \frac{1}{2} g_r h^2 \right) + g_r h\, \partial_y b = 0$
%             }
%         & \shortstack{
%             \(h\): water depth \\
%             \((u,v)\):  velocity \\
%             \(b(x,y)\): bathymetry
%             }
%         & \(\bigl(h(x,y,0),\,b(x,y)\bigr)\)
%         & Solution \(\bigl(h,u,v\bigr)\)
%         & \shortstack{
%             Space: \(128\times128\) \\
%             Time: \(101\) steps
%             } \\
%         \hline
%          &  &  &  &  & \\
%          \hline
%          &  &  &  &  & \\
%          \hline
%          &  &  &  &  & \\
%          \hline
%          &  &  &  &  & \\
%          \hline
%     \end{tabular}
%     \caption{Caption}
%     \label{tab:my_label}
% \end{table*}

\begin{table*}[ht!]
\scriptsize
\centering
\caption{Hyperparameters for each dataset}
\label{tab:hyperparameters}
\begin{tabular}{|l|c|c|c|c|c|}
\hline
% \textbf{Dataset}
  & \makecell{\textbf{Rank \(L\)}} 
  & \makecell{\textbf{Lifting} \textbf{Dimension} \\ \(\boldsymbol{z\!\to\!v}\)} 
  & \makecell{\textbf{Singular Net}\\ \textbf{Type}}
  & \makecell{\textbf{Number of}\\\textbf{Singular Net Layers}} 
  & \makecell{\textbf{Hidden-}\\\textbf{Layer Dimension}} \\
\hline
Shallow Water       & 4 & 512 & MLP  & 3 & 64  \\
Allen Cahn          & 8 & 128 & LSTM & 4 & 32  \\
Diffusion Sorption  & 8 & 128 & MLP  & 6 & 32  \\
Diffusion Reaction  & 3 & 512 & LSTM & 3 & 128 \\
Darcy Flow          & 9 &  128 & MLP  & 2 & 128  \\
\hline
\end{tabular}
\end{table*}

\begin{table*}[ht!]
\scriptsize
\centering
\caption{Mean $L_2$ relative error, expressed as a percent.
Best results are in \textbf{bold} and second-best are \underline{underlined}.
The $\pm$ values denote 95\% Confidence Intervals. All improvements are statistically significant at the 0.05 level (paired t-test, see Appendix).
(N/A indicates a value not reported in the original paper.)}
\renewcommand{\arraystretch}{1.15}
\setlength{\tabcolsep}{5pt}
\begin{tabular}{|l|c|c|c|c|c|c|c|}
\hline

  & \textbf{SVD-NO}              
  & \textbf{DeepONet}        
  & \textbf{FNO}             
  & \textbf{U-NO}            
  & \textbf{MPNN}        
  & \textbf{PINO}           
  & \textbf{U-Net}           \\
\hline
\textit{Category} 
& \multicolumn{1}{c|}{\textit{Ours}}
& \multicolumn{1}{c|}{\textit{DeepONet}}
& \multicolumn{1}{c|}{\textit{Fourier}}
& \multicolumn{1}{c|}{\textit{Fourier}}
& \multicolumn{1}{c|}{\textit{Graph}}
&\textit{PINN, Fourier}
& \multicolumn{1}{c|}{\textit{ML}}
\\
\hline
Shallow Water    
  & \textbf{0.37} $\pm$  0.042              
  & 1.11 $\pm$ 0.235          
  & 0.49 $\pm$  0.022                  
  & 0.52 $\pm$  0.042                  
  & 0.50 $\pm$ 0.031         
  & \underline{0.46} $\pm$ 0.002                     
  & 10.93  $\pm$ 0.706                            \\
% \hline
Allen Cahn  
  & \textbf{0.06}  $\pm$ 0.007                   
  & 16.53 $\pm$  0.230 
  & \underline{0.08} $\pm$  0.001  
  & 0.30 $\pm$ 0.013                    
  & 0.33  $\pm$ 0.009                   
  & \underline{0.08}  $\pm$  0.001      
  & 68.93  $\pm$  0.954                           \\
% \hline
Diffusion Sorption  
  & \textbf{0.10} $\pm$ 0.002          
  & \underline{0.11}  $\pm$ 0.001                  
  & \underline{0.11} $\pm$ 0.001                       
  & \underline{0.11}  $\pm$ 0.001                      
  & 0.29 $\pm$ 0.007                  
  & \underline{0.11}  $\pm$ 0.001                      
  & 12.24    $\pm$ 0.002                 \\
% \hline
Diffusion Reaction
  & \textbf{0.33} $\pm$ 0.010             
  & 0.61  $\pm$ 0.001                                   
  & \underline{0.39}$\pm$ 0.014         
  & 0.40    $\pm$ 0.001                 
  & \underline{0.39} $\pm$ 0.015        
  & 0.43  $\pm$    0.014                
  & 6.70 $\pm$  0.053                   \\
% \hline
Darcy Flow         
  & 2.55 $\pm$ 0.030                    
  & 4.91 $\pm$ 0.116                     
  & \underline{2.35} $\pm$ 0.023        
  & \textbf{2.02} $\pm$  0.028          
  & {\tiny\textsc{N/A}}
  & 3.40 $\pm$ 0.071                    
  & 50.20 $\pm$  0.017                    
\\
\hline
\end{tabular}
\label{tab:results}
\end{table*}

\subsection{Datasets}
\label{sec:datasets}

\paragraph{2D Shallow Water.}  
Simulates fluid flow under gravity over variable topography: \\
$\partial_t h + \partial_x(hu) + \partial_y(hv) = 0$ \\
$\partial_t (hu) + \partial_x \left( u^2 h + \frac{1}{2} g_r h^2 \right) + g_r h\, \partial_x b = 0$ \\
$\partial_t (hv) + \partial_y \left( v^2 h + \frac{1}{2} g_r h^2 \right) + g_r h\, \partial_y b = 0$ \\
%where \(h\) is water depth, \((u,v)\) velocity, and \(b(x,y)\) bathymetry.  \\
%Grid: \(128\times128\) spatial, \(101\) time steps. Samples: 1{,}000. \\
Input: \(\bigl(h(x,y,0),\,b(x,y)\bigr)\) at \(128^2\) points. \\
Output: Solution \(\bigl(h,u,v\bigr)\) on the full \(128\times128\times101\) grid.

\paragraph{1D Allen-Cahn.}  
Models phase separation: \\
\(
  \partial_t u - \epsilon\,\partial_{xx}u + 5u^3 - 5u = 0,\;\epsilon=10^{-4}
  \quad\text{on }(-1,1)\times(0,1].
\) \\
%Grid: \(1024\) spatial (subsampled to 256 points), \(101\) time steps.  
%Samples: 10{,}000.  \\
Input: Initial conditions \(u(x,0)\in\mathbb{R}^{256}\).  \\
Output: Full solution \(u(x,t)\in\mathbb{R}^{256\times101}\).

\paragraph{1D Diffusion-Reaction.}  
Models diffusion with logistic‐type reaction: \\
\(
  \partial_t u - 0.5\,\partial_{xx}u - u(1-u) = 0
  \quad\text{on }(0,1)\times(0,1]\;( \text{periodic} ).
\) \\
%Grid: \(1024\) spatial (subsampled to 256 points), \(101\) time steps.  
%Samples: 10{,}000.  \\
Input: Initial conditions \(u(x,0)\in\mathbb{R}^{256}\).  \\
Output: Full solution \(u(x,t)\in\mathbb{R}^{256\times101}\).

\paragraph{1D Diffusion-Sorption.}  
Models solute transport with nonlinear sorption: \\
\(\partial_t u - \frac{D}{R(u)}\,\partial_{xx}u = 0\) on \((0,1)\times(0,1]\) \\
Boundary conditions: \(u(0,t)=1,\;u_x(1,t)=D^{-1}u(1,t) \) \\
with $D=5\times10^{-4}$ and Freundlich $R(u)$.  \\
%Grid: \(1024\) spatial (subsampled to 256 points), \(201\) time steps.  
%Samples: 10{,}000.  \\
%Input: Initial conditions \(u(x,0)\sim\mathcal U(0,0.2)\in\mathbb{R}^{256}\).  \\
Input: Initial conditions \(u(x,0)\in\mathbb{R}^{256}\).  \\
Output: Full solution \(u(x,t)\in\mathbb{R}^{256\times201}\).

% \paragraph{1D Allen--Cahn.} Models phase separation:
% $\partial_t u-\epsilon\,\partial_{xx}u+5u^3-5u=0$, $\epsilon=10^{-4}$ on $(-1,1)\times(0,1]$.  
% Grid: \(1024\) spatial, \(101\) time steps. Samples: 10{,}000.
% Input: Initial field \(u(x,0)\in\mathbb{R}^{1024}\).  
% Output: Trajectory \(u(x,t)\in\mathbb{R}^{1024\times101}\).  

% \paragraph{1D Diffusion--Reaction.} Diffusion with
% logistic-type reaction:
% $\partial_t u-0.5\,\partial_{xx}u-u(1-u)=0$ on $(0,1)\times(0,1]$ (periodic).   
% Grid: \(1024\) spatial, \(101\) time steps. Samples: 10{,}000.  
% Input: \(u(x,0)\in\mathbb{R}^{1024}\).  
% Output: \(u(x,t)\in\mathbb{R}^{1024\times101}\).

% \paragraph{1D Diffusion--Sorption.} Models solute transport in porous media with nonlinear sorption:
% $\partial_t u-\tfrac{D}{R(u)}\,\partial_{xx}u=0$, $D=5\!\times\!10^{-4}$; Freundlich $R(u)$.  
% $u(0,t)=1$, $u_x(1,t)=D^{-1}u(1,t)$.   
% Grid: \(1024\) spatial, \(201\) time steps.  Samples: 10{,}000
% Input: \(u(x,0)\sim\mathcal U(0,0.2)\in\mathbb{R}^{1024}\).  
% Output: \(u(x,t)\in\mathbb{R}^{1024\times101}\).

\paragraph{2D Darcy Flow.} 
Models steady incompressible flow in porous media: \\
$-\nabla\!\cdot(a\nabla u)=1$ on $(0,1)^2$ \quad $u|_{\partial\Omega}=0$ \\
%with binary permeability $a\!\in\!\{0.1,1\}^{128\times128}$.  \\
%Grid: \(128\times128\).  
%Samples: 10{,}000.    \\
Input: Permeability field \(a(x,y)\).  \\
Output: Solution \(u(x,y)\in\mathbb{R}^{128\times128}\).

% \paragraph{1D Burgers’.} Models nonlinear advection–diffusion in fluids:
% $\partial_t u+\tfrac12\partial_x(u^2)=\nu\,\partial_{xx}u$, $\nu=0.1$ on $(0,1)\times(0,1]$ (periodic).  
% Model maps $u(x,0)\!\rightarrow\!u(x,1)$ on a 1024–point grid.  
% $1{,}000/100$ train/test samples.

\vspace{0.5em}
\noindent
Together these five datasets cover elliptic (Darcy), reaction–diffusion (Diffusion--Reaction/Sorption, Allen–Cahn) and fluid‐dynamics (Shallow Water) regimes, providing a broad test bed for neural operators.

\subsection{Experimental Setup}

The experiments were conducted on an NVIDIA L40 49GB GPU. Following the PDENNEval protocol \cite{pdenneval}, we trained for 500 epochs on each PDE benchmark (200 epochs for Shallow Water), and split the data into 10\% test, 10\% validation, and 80\% training sets. Our training loss is:
\(
\mathcal{L}_{\text{total}} = L_2 + \mathcal{L}_{\text{orth}},
\)
(Eqs.~\ref{eq:mean_l2_rel_err}, \ref{eq:orth_loss}), where $L_2$ is the reconstruction error (without the $100$ pre-factor) and $\mathcal{L}_{\text{orth}}$ enforces singular‐function orthonormality. 
Moreover, all time-steps were learned jointly as in \cite{fno}. Hyperparameter tuning was performed on the validation dataset. The \textsc{Adam} optimizer was used, with an initial learning rate of $10^{-3}$.

\paragraph{Hyperparameter Tuning.}
The hyperparameters used are provided in Table~\ref{tab:hyperparameters}. These include configurations for the SVD-NO model, such as the rank \(L\) of the singular functions, the lifting dimension (from \(z\) to \(v\)), the type of singular net used, number of singular net layers, and their hidden layer dimension. The activation function $\gamma$ in the SVD-NO layer was Gelu, and the number of SVD-NO layers was 4.

\paragraph{Baseline Hyperparameters.}
For all baselines, we used the hyperparameter settings provided in \cite{pdenneval}.

\section{Results}
\label{sec:results}
\subsection{Main Results}

The experimental results, summarized in Table~\ref{tab:results}, demonstrate the performance of SVD-NO compared to other SOTA methods across various PDE benchmarks. The Mean $L_2$ Relative Error ($\%$) was used as the performance metric. The model achieves SOTA performance and significant $L_2$ reductions, with average improvement percentages of: $17.8\%$ in Shallow-Water, $25.0\%$ in Allen-Cahn, $9.1\%$ in Diffusion-Sorption, and, $15.4\%$ in Diffusion-Reaction. The improvement percentage was calculated by comparing the model to the best-performing baseline per PDE.  In terms of the one remaining equation, Darcy Flow, SVD-NO places third.

In addition, SVD-No achieves $L_\infty$ reductions, with improvement percentages up to $26.80\%$. The full $L_\infty$ results can be found in the appendix.

% The radar plot in Figure~\ref{fig:spider_plot} further illustrates the comparative performance of each model across the benchmarks. As seen, the SVD-NO model (in blue) outperforms other methods in several benchmarks. While models like U-NO and MPNN perform well on some datasets, SVD-NO exhibits a more balanced performance across all benchmarks.

\paragraph{Significance of Results.}
To assess robustness, each experiment was repeated ten times with different random seeds. For each setting we report the sample mean and its 95\% confidence interval (CI). These statistics are presented in Table \ref{tab:results}.
Moreover, we conducted paired t-tests on datasets where SVD-NO performed best, yielding \textit{p-values} well below 0.05. Full results are provided in the appendix.

% \textcolor{blue}{Moreover, we performed paired t-tests across datasets where SVD-NO
% achieved the lowest result. In nearly all cases, the p-values are well below the 0.05 threshold, indicating that SVD-NO’s improvements are statistically significant rather than due to random chance. The full p-value results can be found in the appendix.}

% \begin{table}[h!]
% \centering
% \caption{Mean $L_2$ results (± 95\% confidence interval)}
% \label{tab:ci}
% \begin{tabular}{|l|c|}
% \hline
% \textbf{PDE}               & \textbf{Confidence interval}      \\
% \hline
% Shallow Water              & $0.33 \pm 0.125$                  \\
% Allen Cahn                 & $0.04 \pm 0.085$                  \\
% Diffusion Sorption         & $0.10 \pm 0.001$                  \\
% Diffusion Reaction         & $0.36 \pm 0.021$                  \\
% Darcy Flow                 & $3.06 \pm 0.030$                  \\
% \hline
% \end{tabular}

% \vspace{0.5em}
% \begin{minipage}{0.95\linewidth}
% \end{minipage}
% \end{table}

\paragraph{Orthogonality.} We assess the orthogonality of the learned basis functions. The values of $\mathcal{L}_{\text{ortho}}$ range from $1.4 \times 10^{-7}$ for Diffusion-Sorption to $3.4 \times 10^{-5}$ for Darcy Flow.  The training is thus quite effective at simultaneously orthogonalizing while learning to solve the PDEs.

% The results in Table~\ref{tab:ortho} show the orthogonality loss, $\mathcal{L}_{\text{ortho}}$, achieved on each benchmark. Lower values indicate closer adherence to an orthonormal basis.

% \begin{table}[ht!]
% \scriptsize
% \centering
% \caption{Orthogonality loss $\mathcal{L}_{\text{ortho}}$. Lower values indicate closer adherence to an orthonormal basis.}
% \label{tab:ortho}
% \begin{tabular}{|l|c|}
% \hline
% % \textbf{PDE}               & \textbf{$\mathcal{L}_{\text{ortho}}$} \\
% & \textbf{$\mathcal{L}_{\text{ortho}}$} \\
% \hline
% Shallow Water              & 4.5e-6  \\
% Allen Cahn                 & 2.3e-5  \\
% Diffusion Sorption         & 1.4e-7  \\
% Diffusion Reaction         & 1.8e-5  \\
% Darcy Flow                 & 3.4e-5  \\
% \hline
% \end{tabular}

% \vspace{0.5em}
% \begin{minipage}{0.95\linewidth}
% \end{minipage}
% \end{table}

\paragraph{Performance vs. Solution Spatial Variability.}
We hypothesize that SVD-NO will give greater performance gains when the solution is more challenging.  While it is not straightforward to quantify how challenging a solution is, we may use a simple heuristic: how much spatial variation there is in the solution.  In particular, if the solution is $u(x)$ we may examine the following quantity:
\[
\beta(h)
\;=\;
\frac{\int (u(x+h) - u(x))^2 dx}{\mathrm{Var}(u)}
\]
where $h \in \mathbb{R}^{d_x}$ is an offset, and $\mathrm{Var}(u)$ is the variance of $u$.  This quantity (which is related to the ``Variogram'' \cite{matheron1963principles})  measures how similar the solution is when comparing any point $x$ to a point that is $h$ away from $x$, and is normalized by $\mathrm{Var}(u)$ to remove scaling effects.  In our case, we will compute a single number by averaging over many possible offsets, 
\[
\beta = \frac{1}{|\mathcal{H}|} \sum_{h \in \mathcal{H}} \beta(h)
\]
In practice, we take $\mathcal{H} = \{1, \dots, 5\}^{d_x}$: we allow grid points which are up to 5 away in any spatial direction.  Finally, if the PDE solution depends on time, then we will compute $\beta$ for each time $t$, and then average over the results.

% \paragraph{Stationarity.}
% A random field is \emph{second‐order stationary (weak stationarity)} if its mean is constant and its covariance depends only on the separation (lag) between two points, not on their absolute locations \cite{cressie1993statistics}. To quantify how “stationary” each dataset is, we use the \emph{normalized spatial variogram}:
% \[
% \frac{\gamma(h)}{\mathrm{Var}(f)}
% \;=\;
% \frac{1}{2\,\mathrm{Var}(f)}\,
% \mathbb{E}\bigl[\bigl(f(x)-f(x+h)\bigr)^2\bigr],
% \]
% where \(\gamma(h)\) is the semivariogram measuring the average squared difference between points separated by lag \(h\) \cite{matheron1963principles} and $\mathrm{Var}(f)$ is the variance of $f$. We compute the variogram for lags \(h = 1,\dots,5\) along each spatial dimension.  We then average \(\gamma(h)/\mathrm{Var}(f)\) over those lags and all data points. Intuitively, a \emph{small} normalized variogram means nearby points are very similar—high spatial stationarity, whereas a \emph{large} value indicates stronger spatial inhomogeneity.% \cite{isaaks1989applied}.

\begin{figure}[t]
  \centering
  \begin{tikzpicture}
    \begin{axis}[
      width=0.9\columnwidth, height=0.5\columnwidth,
      font=\tiny,
      title={\textbf{Percentage Improvement vs.\ Solution Spatial Variability
      }},
      xlabel={Solution Spatial Variability ($\beta$)},
      ylabel={Improvement (\%)},
      ylabel style={yshift=-5pt},
      xmin=0, xmax=0.11,
      ymin=-5, ymax=30,
      legend style={
        at={(0.02,0.98)}, anchor=north west,
        draw=none, font=\tiny
      },
      every axis plot/.append style={
        mark=*, mark size=1.5pt
      },
      tick label style={font=\tiny},
    ]

      % raw data
      \addplot+[only marks, blue] coordinates {
        (0.0174,  0.00)    % Darcy Flow
        (0.0293,  15.38)    % Reaction Diffusion
        (0.0019,  9.09)    % Reaction Sorption
        (0.1027, 25.00)    % Allen Cahn
        (0.0954, 19.78)    % Shallow Water
      };

      % fit line
      \addplot+[orange, domain=0.001:0.105, mark=none, samples=2, thick]
        {174*x + 5};
      %\addlegendentry{fit: \(y=163.6*x + 5\)}

      % point labels
      \node[anchor=west]       at (axis cs:0.0174,  0.00) {Darcy Flow};
      \node[anchor=west]       at (axis cs:0.0293,  15.38) {Reaction Diffusion};
      \node[anchor=west] at (axis cs:0.0019,  9.09) {Reaction Sorption};
      \node[anchor=east] at (axis cs:0.1027, 25.00) {Allen Cahn};
      \node[anchor=east] at (axis cs:0.0954, 19.78) {Shallow Water};

    \end{axis}
  \end{tikzpicture}
  \caption{
  SVD-NO’s percentage improvement over the best baseline versus the solution's spatial variability ($\beta$) for each PDE. Each point represents a dataset; the least‐squares fit shows that more challenging problems (higher solution spatial variability) correlate with larger improvements.
}
  \label{fig:spatial-stationarity}
\end{figure}
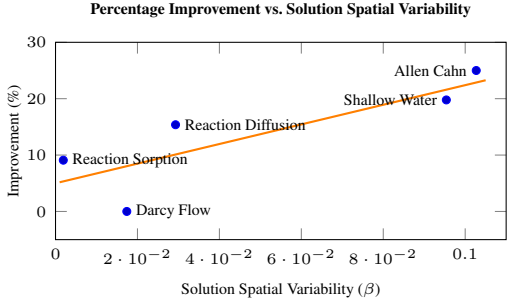

Figure~\ref{fig:spatial-stationarity} shows, for each PDE benchmark, the spatial variability of its solution as measured by $\beta$ on the horizontal axis, vs. ~SVD-NO's percent improvement over the best baseline on the vertical axis.  The least‐squares fit reveals a clear positive trend: PDEs whose solutions are characterized by higher spatial variability yield larger performance gains.  In other words, our method delivers the greatest improvements on more spatially inhomogeneous problems, while on problems with smoother solutions the benefits are smaller.  Thus, this trend serves to illustrate that the more expressive kernel provided by SVD-NO helps the most on problems with more challenging solutions.

\begin{figure}[htbp]
  \centering
  \begin{tikzpicture}
    \begin{groupplot}[
      group style={
        group size=2 by 1,
        horizontal sep=1cm,
      },
      width=4.8cm,
      height=3.5cm,
      tick label style={font=\tiny},
      title style={font=\tiny},
      xlabel={Number of singular functions \(L\)},
      ymin=-0.1, ymax=1.1,
      xtick={3,4,5,6,7,8,9,10},
      ylabel style={yshift=-5pt,font=\tiny},
      xlabel style={yshift=5pt,font=\tiny},
      legend style={
        at={(-0.8,-0.3)}, anchor=north west,
        draw=none, font=\tiny,
         mark=*,
        mark options={solid,scale=0.1},
        only marks
      },
    % legend to name=globalLegend,
      legend columns=3,
      % legend style={
      %   /tikz/column 2/.style={column sep=1cm},
      % }
    ]

    % Left: normalized test loss
    \nextgroupplot[
      ylabel={Normalized test loss},
      title={\textbf{Test Loss}}
    ]
    \addplot[blue, thick] table[x=L, y expr={(\thisrow{loss}-0.046523)/(0.100000-0.046523)}] {
L loss
3   0.092887
4   0.100000
5   0.055831
6   0.060352
7   0.068482
8   0.046523
9   0.064950
10  0.060000
    }; 
    % \addlegendentry{Allen–Cahn}

    \addplot[red, thick] table[x=L, y expr={(\thisrow{loss}-0.107756669)/(0.112866089-0.107756669)}] {
L loss
3   0.112866089
4   0.110711929
5   0.108350205
6   0.109285655
7   0.107862392
8   0.107756669
9   0.109059036
10  0.108574010
    };
    % \addlegendentry{Diffusion–Sorption}

    \addplot[teal, thick] table[x=L, y expr={(\thisrow{loss}-0.374843415)/(0.505102340-0.374843415)}] {
L loss
3   0.442138568
4   0.505102340
6   0.412120618
7   0.431144867
8   0.424831305
9   0.426620112
10  0.374843415
    }; 
    % \addlegendentry{Shallow Water}

    % Right: normalized GPU memory
    \nextgroupplot[
      ylabel={Normalized GPU memory},
      title={\textbf{GPU Memory}}
    ]
    % Allen–Cahn mem
    \addplot[blue, thick] table[x=L, y expr={(\thisrow{mem}-841)/(2215-841)}] {
L mem
3 850
4 1020
5 1210
6 1410
7 1610
8 1820
9 2020
10 2220
    }; 
    % \addlegendentry{Allen–Cahn}

    % Diffusion–Sorption mem
    \addplot[red, thick] table[x=L, y expr={(\thisrow{mem}-803)/(2177-803)}] {
L mem
3 800
4 970
5 1160
6 1360
7 1570
8 1770
9 1970
10 2170
    }; 
    % \addlegendentry{Diffusion–Sorption}

    % Shallow Water mem
    \addplot[teal, thick] table[x=L, y expr={(\thisrow{mem}-19086)/(54230-19086)}] {
L mem
3 19086
4 23566
5 28625
6 32000
7 38867
8 42483
9 52620
10 54230
    }; 
    % \addlegendentry{Shallow Water}

\addlegendentry{Allen–Cahn}
\addlegendentry{Diffusion–Sorption}
\addlegendentry{Shallow Water}
    \end{groupplot}
  \end{tikzpicture}
\caption{(Left) Min–max normalized test‐loss curves and (Right) min–max normalized peak GPU memory usage vs.\ SVD rank \(L\).  Increasing \(L\) yields significant accuracy gains but also linearly higher memory, illustrating the practical trade‐off between model fidelity and resource consumption.}

  \label{fig:loss-and-mem}
\end{figure}
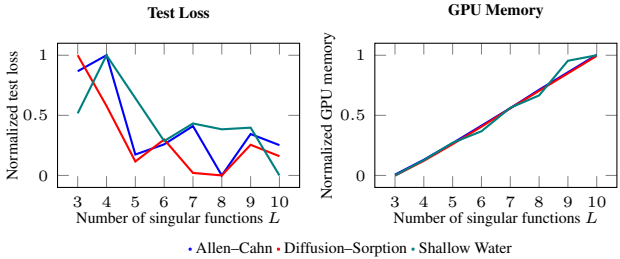

\paragraph{Memory–Accuracy Trade‐off.}
Figure~\ref{fig:loss-and-mem} illustrates that increasing the SVD rank \(L\) steadily reduces the normalized test loss (left panel), at the cost of higher GPU memory usage (right panel).  In particular, moving from \(L=3\) to \(L=10\) (well below the full data dimensionality) substantially improves accuracy, while peak memory requirements grow linearly (as expected from the \(O(n\,d\,L)\) memory complexity). This clear memory–accuracy trade‐off allows practitioners to pick \(L\) based on their hardware budget. \\
A similar linear trend in runtime, proportional to \(O(n\,d\,L)\), was also observed, but the plot is omitted for brevity. 
\\
To enable overlaying curves from different datasets on the same scale, each dataset’s raw test losses and peak GPU memory were min–max normalized.

% \paragraph{\textcolor{blue}{Impact of SVD Rank $L$ on Test Loss.}}
% In Figure~\ref{fig:rank-vs-loss} we observe a clear dependence of the generalization error on the chosen SVD rank \(L\) (number of singular values).  As we vary \(L\) from 3 up to 10 (well below the full data dimensionality), the test loss drops sharply, illustrating that retaining more modes substantially improves accuracy.

\subsection{Ablation Studies}
\label{sec:ablation}
% To disentangle the contribution of each design choice in SVD-NO we conduct the following ablations. Results in Table~\ref{tab:ablations}

To evaluate the impact of each design choice in SVD-NO, we conduct the following ablations, with results in Table \ref{tab:ablations}.

\paragraph{Direct MLP Kernel}
To isolate the contribution of the low-rank SVD, we replace the factorized kernel with a fully connected network that learns a mapping directly, i.e.~
\[
\tilde{\kappa}_\theta(z,z') \;=\; \mathrm{MLP}(z,z') .
\]
Removing the rank-\(L\) constraint substantially degrades accuracy: the mean relative error rises by $3.02\times$ across benchmarks. The dense formulation is also markedly slower, the average training time per epoch increases from \textit{17.64}s to \textit{85.73}s on Diffusion–Reaction, from \textit{2.32}s to \textit{176.19}s on Diffusion–Sorption, from \textit{4.87}s to \textit{74.31}s on Allen–Cahn, and from \textit{22.36}s to \textit{374.75}s on Shallow–Water. These results underscore the dual advantage of SVD-NO: it achieves higher accuracy while reducing the per-layer complexity from \(O(n^{2}d^2)\) for a generic dense kernel to \(O(ndL)\).

\paragraph{Mercer Theorem.}
To measure the benefit of the SVD, we swap it for Mercer's Theorem \cite{mercer}, valid for continuous, symmetric, positive-definite kernels.  
We write
\(
\kappa(z,z')=\sum^{\infty}_{\ell=1}\lambda_\ell\,\phi_\ell(z)\,\phi_\ell(z'),
\)
with $\lambda_\ell \ge 0$, and keep only the first \(L\) terms.  The truncated eigenpairs
\(\{\lambda_\ell,\phi_\ell\}_{\ell=1}^{L}\) are learned end-to-end, giving an update analogous to Eq (\ref{eq:update_continuous}).
% the update
% \[
% v^{t+1}(z)=\sum_{\ell=1}^{L}\lambda_\ell\,\phi_\ell(z)\!
%            \int_{\mathcal Z}\phi_\ell(z')\,v^{t}(z')\,\mathrm d\mu(z').
% \]
For vector-valued fields, we use the block form
\(\kappa(z,z')=\Phi(z)\,\Lambda\,\Phi(z')^{\!\top}\) with
\(\Lambda=\mathrm{diag}(\lambda_1,\dots,\lambda_L)\) as in \cite{mercer_extension}.

Replacing SVD with the Mercer expansion degrades performance, increasing the mean error by \(3.32\times\) across datasets. This drop is expected: the Mercer theorem applies only to \emph{symmetric, positive-definite} kernels, whereas the SVD factorisation holds for any Hilbert–Schmidt kernel operator, making SVD-NO applicable in a broader range.

% \begin{table}[t!]

% \scriptsize
%     \centering
%     \caption{Mean $L_2$ relative error (\%) for each ablation}
%     \label{tab:ablations}
%     \begin{tabular}{|l|c|c|c|c|}
%         \hline
%         \textbf{Ablation}
%       & \makecell{\textbf{Allen}\\\textbf{Cahn}}
%       & \makecell{\textbf{Diffusion}\\\textbf{Sorption}}
%       & \makecell{\textbf{Diffusion}\\\textbf{Reaction}}& \makecell{\textbf{Shallow}\\\textbf{Water}}\\
%         \hline
%         SVD-NO (full)          & \textbf{0.06} & \textbf{0.10} & \textbf{0.33} &\textbf{0.37}\\
%         Direct\,MLP\,Kernel    & 0.49          & 0.11          & 0.88          & 0.99 \\        
%         Mercer                 & 0.99          & 0.13          & 0.54          & 0.87\\
%         No Orthogonality loss               & 0.84          & 0.11          & 0.51         & 0.76\\
%         \hline
%     \end{tabular}
% \end{table}

\begin{table}[t!]
\scriptsize
\centering
\caption{Mean $L_2$ relative error (\%) for each dataset across ablations}
\label{tab:ablations}
\begin{tabular}{|l|c|c|c|c|}  
    \hline
    \textit{Ablation} 
    & \textbf{\makecell{SVD-NO}}
    & \textbf{\makecell{Direct MLP\\Kernel}} 
    & \textbf{Mercer} 
    & \textbf{\makecell{No $\mathcal{L}_{\text{ortho}}$ \\Penalty}} \\
    \hline
    
    Shallow Water 
    & \textbf{0.37} 
    & 0.99 
    & 0.87 
    & 0.76 \\
    Allen Cahn
    & \textbf{0.06} 
    & 0.49 
    & 0.99 
    & 0.84 \\
    
    Diffusion Sorption
    & \textbf{0.10} 
    & 0.11 
    & 0.13 
    & 0.11 \\
    
    Diffusion Reaction
    & \textbf{0.33} 
    & 0.88 
    & 0.54 
    & 0.51 \\
    \hline
\end{tabular}
\end{table}

\paragraph{No Orthogonality Penalty.}
We assess the impact of the orthogonality regulariser by training SVD-NO using only the
$L_2$ reconstruction loss, i.e.\ setting $\mathcal L_{\text{ortho}}\!=\!0$.  
% Comparing against the full model quantifies how much the Gram-matrix penalty contributes to stability and accuracy.
Eliminating the Gram–matrix penalty markedly degrades performance: the mean relative error rises by \(2.97\times\) across benchmarks. Without the orthogonality constraint the learned set of singular functions \(\Phi\) and \(\Psi\) drift away from an (approximate) orthonormal basis, so the factorisation no longer behaves as an SVD, reducing predictive accuracy.
%\\
%\\
%\emph{Remark:} In the Allen–Cahn equation, the cubic nonlinearity, $5u^3-5u$, drives the field into two nearly constant phases separated by very sharp, nonlinear interfaces and hence cannot be well captured by symmetric Mercer expansions or by non-orthogonal singular function nets.

\section{Conclusions}
SVD-NO is a new neural operator framework which allow for the learning of highly expressive kernels, bridging Hilbert–Schmidt theory and deep learning.  By learning the kernel’s left and right singular functions and singular values end-to-end, SVD-NO provides a low-rank factorization that applies to any square-integrable kernel and scales linearly with the spatial resolution.  A Gram-matrix orthogonality penalty preserves the SVD structure during training.  Comprehensive experiments on five diverse PDEs demonstrate that SVD-NO sets a new state of the art reducing the mean $L_2$ error of the best competing models by up to 25\%. Ablation studies further verify the importance of the SVD parameterization and the orthogonality regularizer.

\bibliography{aaai2026}

\end{document}